\documentclass[twoside,11pt]{article} 
\usepackage{jmlr2e} 

\jmlrheading{15}{2014}{141-145}{4/13; Revised 10/13}{1/14}{Marc Claesen, Frank De Smet, Johan A.K. Suykens and Bart De Moor}

\ShortHeadings{EnsembleSVM}{Claesen, De Smet, Suykens and De Moor}
\firstpageno{1}

\usepackage{amsmath}
\usepackage{tikz}
\usepackage{ifpdf}
\usepackage{multirow}
\usepackage{afterpage}
\usepackage{booktabs}
\usepackage{fancyvrb}
\usepackage{color}
\usepackage[colorlinks=false,allbordercolors={1 1 1}]{hyperref}

\newcommand{\esvm}{\texttt{\selectfont EnsembleSVM}}
\newcommand{\esvmshort}{\texttt{ESVM}}
\newcommand{\libsvm}{\texttt{LIBSVM}}

\begin{document}

\title{\esvm: A Library for Ensemble Learning Using Support Vector Machines}

\author{\name Marc Claesen \email marc.claesen@esat.kuleuven.be  \\ 
\addr KU Leuven, ESAT -- STADIUS/iMinds Future Health  \\
Kasteelpark Arenberg 10, box 2446 \\
3001 Leuven, Belgium \AND
\name Frank De Smet \email frank.desmet@cm.be\\ 
\addr KU Leuven, Department of Public Health and Primary Care, Environment and Health \\
Kapucijnenvoer 35 blok d, box 7001 \\
3000 Leuven, Belgium \AND
\name Johan A.K. Suykens \email johan.suykens@esat.kuleuven.be   \\ 
\name Bart De Moor \email bart.demoor@esat.kuleuven.be  \\ 
\addr KU Leuven, ESAT -- STADIUS/iMinds Future Health \\
Kasteelpark Arenberg 10, box 2446 \\
3001 Leuven, Belgium
}

\editor{Geoff Holmes}

\maketitle

\begin{abstract}
\esvm\ is a free software package containing efficient routines to perform
ensemble learning with support vector machine (SVM) base models. It currently
offers ensemble methods based on binary SVM models. Our implementation
avoids duplicate storage and evaluation of support vectors which are shared between constituent models. Experimental results show that using
ensemble approaches can drastically reduce training complexity while maintaining
high predictive accuracy. The \esvm\ software package is freely available online
at \texttt{\url{http://esat.kuleuven.be/stadius/ensemblesvm}}.
\end{abstract}


\section{Introduction}

Data sets are becoming increasingly large. Machine learning practitioners are confronted with
problems where the main computational constraint is the amount of time available. Problems become particularly challenging when
the training sets no longer fit into memory. Accurately solving the dual problem
for SVM training with nonlinear kernels requires a run time which is at least
quadratic in the size of the training set $n$, thus training complexity is $\Omega(n^2)$
\citep{bottou07svm,NL09a}.

\esvm\ employs a divide-and-conquer strategy by aggregating many SVM models,
trained on small subsamples of the training set. 
Through subdivision, total training time decreases significantly, even though
more models need to be trained. For example, training $p$
classifiers on subsamples of size $n/p$, results in an approximate complexity
of $\Omega(n^2/p)$. This reduction in complexity helps in dealing
with large data sets and nonlinear kernels.

Ensembles of SVM models have been used in various applications
\citep{Wang20096466,5646323,citeulike:9860864}.
\citet{Collobert02} use ensembles for large scale learning and employ a neural
network to aggregate base models.
\citet{Valentini03lowbias} provide an implementation which allows base models to use different kernels.
For efficiency reasons, we require base models to share a single kernel function. 

While other implementations mainly focus on improving predictive performance, our framework
primarily aims to (i) make nonlinear large-scale learning feasible through complexity reductions and (ii) enable fast prototyping of novel ensemble algorithms.

\section{Software Description}
The \esvm\ software is freely available online under a
LGPL license. \esvm\ provides ensembles of
instance-weighted SVMs, as defined in Equation~\eqref{weightedsvm}.
The default approach we offer is bagging, which is commonly used to improve the
performance of unstable classifiers \citep{Breiman:1996:BP:231986.231989}. In
bagging, base models are trained on bootstrap subsamples of the training set
and their predictions are aggregated through majority voting.

Base model flexibility is maximized by using instance-weighted binary support
vector machine classifiers, as defined in Equation~\eqref{weightedsvm}. This formulation lets users define
misclassification penalties per training instance $C_i,\ i=1,\ldots,n$ and
encompasses popular approaches such as C-SVC and class-weighted SVM
\citep{Cortes:1995:SN:218919.218929, osuna1997}.
\begin{align}
\mathbf{\min_{\mathbf{w},\mathbf{\xi},\rho} } \ &
\frac{1}{2}\mathbf{w}^T\mathbf{w}+\sum_{i=1}^n C_i \xi_i,
\label{weightedsvm} \\
\text{subject\ to }  &y_i(\mathbf{w}^T\phi(\mathbf{x}_i)+\rho)\geq 1-\xi_i,
&i=1,\ldots,n, \nonumber \\
&\xi_i \geq 0, &i=1,\ldots,n. \nonumber  
\end{align}

When aggregating SVM models, the base models often share support
vectors (SVs). The \esvm\ software intelligently caches distinct SVs to ensure
that they are only stored and used for kernel evaluations once. As a result,
\esvm\ models are smaller and faster in prediction than ensemble implementations based on wrappers.

\subsection{Implementation}
\esvm\ has been implemented in \texttt{C++} and makes heavy use of the standard library. The main implementation focus is training speed. We use facilities provided by the \texttt{C++11} standard and thus require a moderately recent compiler, such as \texttt{gcc} $\geq4.7$ or \texttt{clang} $\geq 3.2$. A portable Makefile system based on GNU autotools is used to build \esvm. 

\esvm\ interfaces with \libsvm\ to train base models \citep{CC01a}. Our code must be linked to \libsvm\ but does not depend on a specific version. This allows users to choose the desired version of the \libsvm\ software in the back-end. 

The \esvm\ programming framework is designed to facilitate prototyping of ensemble algorithms using SVM base models.
We particularly provide extensive support to define novel aggregation schemes, should the available options be insufficient. 
Key components are extensively documented and on a broad overview is provided on our wiki.\footnote{The \esvm\ development wiki is available at \url{https://github.com/claesenm/EnsembleSVM/wiki}.} 

The \esvm\ library was built with extensibility and user
contributions in mind.
Major API functions are well documented to lower the threshold
for external development. The executable tools provided with \esvm\ are
essentially wrappers for the library itself. The tools can be considered as use
cases of the main API functions to help developers.

\subsection{Tools}
The main tools in this package are {\tt esvm-train} and {\tt esvm-predict}, used
to train and predict with ensemble models. Both of these are pthread-parallelized. 
Additionally, the {\tt merge-models} tool can be used to merge standard \libsvm\ models into ensembles. 
Finally, {\tt esvm-edit} provides facilities to modify the aggregation scheme used by an ensemble.

\esvm\ includes a variety of extra tools to facilitate basic operations such as
stratified bootstrap sampling, cross-validation, replacing categorical features
by dummy variables, splitting data sets and sparsifying standard data sets. We recommend retaining the original ratio of positives and negatives in the training set when subsampling.

\section{Benchmark Results} \label{bench}
To illustrate the potential of our software, \esvm\ $2.0$ has been benchmarked with respect to \libsvm\
$3.17$. To keep the experiments simple, we use majority voting to aggregate predictions, even though more sophisticated methods are offered. For reference, we also list the best obtained accuracy with a linear model, trained using \texttt{LIBLINEAR} \citep{Fan:2008:LLL:1390681.1442794}. Linear methods are common in large-scale learning due to their speed, but may result in significantly decreased accuracy. This is why scalable nonlinear methods are desirable.

We used two binary classification problems, namely the {\tt covtype} and {\tt ijcnn1} data sets.\footnote{Both data sets are available at \url{http://www.csie.ntu.edu.tw/~cjlin/libsvmtools/datasets/binary.html}.} Both data sets are balanced. Features were always scaled to $[0,1]$. We have
used C-SVC as SVM and base models ($\forall\ i: C_i=C$). Reported numbers are averages of $5$ test runs to ensure reproducibility. We used the RBF kernel, defined by the kernel function
$\kappa(\mathbf{x}_i,\mathbf{x}_j)=e^{-\gamma ||\mathbf{x}_i-\mathbf{x}_j||^2}$.
Optimal parameter selection was done through cross-validation.

The \texttt{covtype} data set is a common classification benchmark featuring
$54$ dimensions \citep{Blackard00covtype}. We randomly sampled balanced training
and test sets of $100,000$ and $40,000$ instances respectively and classified class $2$ versus all others. The \texttt{ijcnn1} data set was used in a machine learning challenge during IJCNN 2001 \citep{prokhorov2001ijcnn}. It contains $35,000$ training instances in $22$ dimensions.

\begin{table}[!t]
\centering
\begin{tabular}{lcccccccc}
\toprule 
{\bfseries data set} & \multicolumn{3}{c}{\bfseries test set accuracy} & 
\multicolumn{2}{c}{\bfseries no. of SVs} & \multicolumn{2}{c}{\bfseries time
(s)}
\\
 & \libsvm & \texttt{LIBLINEAR} & \esvmshort & \libsvm & \esvmshort & \libsvm
 & \esvmshort \\
\midrule 
{\tt covtype} & $0.92$ & $0.76$ & $0.89$ & $26516$ & $50590$ & $728$ & $35$ \\
{\tt ijcnn1} & $0.98$ & $0.92$ & $0.98$ & $3564$ & $7026$ & $9.5$ & $0.3$ \\
\bottomrule
\end{tabular}
\caption{Summary of benchmark results per data set: test set accuracy, number of
support vectors and training time. Accuracies are listed for a single \libsvm\ model, \texttt{LIBLINEAR} model and an ensemble model.}
\label{resultstable}
\end{table}

Results in Table~\ref{resultstable} show several interesting trends. Training \esvm\ models is orders of magnitude faster,
because training SVMs on small subsets significantly reduces complexity. Subsampling induces smaller kernels per base model resulting in lower overall memory use. Due to our parallelized implementation, ensemble models were faster in prediction than \texttt{LIBSVM} models in both experiments despite having twice as many SVs.

The ensembles in these experiments are competitive with a traditional SVM even though we used simple majority voting. For \texttt{covtype}, ensemble accuracy is $3\%$ lower than a single SVM and for \texttt{ijcnn1} the ensemble is marginally better ($0.2\%$). Linear SVM falls far short in terms of accuracy for both experiments, but is trained much faster ($< 2$ seconds).

We obtained good results with very basic aggregation. \citet{Collobert02} illustrated that more sophisticated aggregation methods can improve the predictive performance of ensembles. Others have reported performance improvements over standard SVM for
ensembles using majority voting \citep{Valentini03lowbias,Wang20096466}.
 
\section{Conclusions}
\esvm\ provides users with efficient tools to experiment with ensembles of
SVMs. Experimental results show that training ensemble models is significantly
faster than training standard \libsvm\ models while maintaining competitive predictive accuracy.

Linear methods are frequently applied in large-scale learning, mainly due to their low training complexity. Linear methods are known to have competitive accuracy for high dimensional problems. As our benchmarks showed, the difference in accuracy may be large for low dimensional problems. As such, fast nonlinear methods remain desirable in large-scale learning, particularly for low dimensional tasks with many training instances. Our benchmarks illustrate the potential of the ensemble approaches offered by \esvm.

Ensemble performance may be improved by using more complex aggregation schemes. \esvm\ currently offers various aggregation schemes, both linear and nonlinear. Additionally, it facilitates fast prototyping of novel methods. 

\esvm\ strives to provide high-quality,
user-friendly tools and an intuitive programming framework for ensemble learning with SVM base models. The software will be kept up to date by incorporating promising new methods and ideas when they are presented in the literature. User requests and suggestions are welcome and appreciated.

\acks{Frank De Smet is a member of the medical management department of the National Alliance of Christian Mutualities. Acknowledged funding sources: Marc Claesen (IWT grant number 111065); Research Council KU Leuven: GOA MaNet, CoE SymBioSys; EU: ERC AdG A-DATADRIVE-B. }


\begin{thebibliography}{14}
\providecommand{\natexlab}[1]{#1}
\providecommand{\url}[1]{\texttt{#1}}
\expandafter\ifx\csname urlstyle\endcsname\relax
  \providecommand{\doi}[1]{doi: #1}\else
  \providecommand{\doi}{doi: \begingroup \urlstyle{rm}\Url}\fi

\bibitem[Blackard and Dean(1999)]{Blackard00covtype}
Jock~A. Blackard and Denis~J. Dean.
\newblock {Comparative accuracies of artificial neural networks and
  discriminant analysis in predicting forest cover types from cartographic
  variables}.
\newblock \emph{Computers and Electronics in Agriculture}, 24\penalty0
  (3):\penalty0 131--151, December 1999.

\bibitem[Bottou and Lin(2007)]{bottou07svm}
L\'{e}on Bottou and Chih-Jen Lin.
\newblock {Support Vector Machine Solvers}.
\newblock In L\'{e}on Bottou, Olivier Chapelle, Dennis {DeCoste}, and Jason
  Weston, editors, \emph{Large Scale Kernel Machines}, pages 301--320,
  Cambridge, MA, USA, 2007. MIT Press.

\bibitem[Breiman(1996)]{Breiman:1996:BP:231986.231989}
Leo Breiman.
\newblock Bagging predictors.
\newblock \emph{Machine Learning}, 24\penalty0 (2):\penalty0 123--140, August
  1996.

\bibitem[Chang and Lin(2011)]{CC01a}
Chih-Chung Chang and Chih-Jen Lin.
\newblock {LIBSVM}: A library for support vector machines.
\newblock \emph{ACM Transactions on Intelligent Systems and Technology},
  2:\penalty0 27:1--27:27, 2011.
\newblock Software available at \url{http://www.csie.ntu.edu.tw/~cjlin/libsvm}.

\bibitem[Collobert et~al.(2002)Collobert, Bengio, and Bengio]{Collobert02}
Ronan Collobert, Samy Bengio, and Yoshua Bengio.
\newblock A parallel mixture of {SVMs} for very large scale problems.
\newblock \emph{Neural Computation}, 14\penalty0 (5):\penalty0 1105--1114,
  2002.

\bibitem[Cortes and Vapnik(1995)]{Cortes:1995:SN:218919.218929}
Corinna Cortes and Vladimir Vapnik.
\newblock Support-vector networks.
\newblock \emph{Machine Learning}, 20\penalty0 (3):\penalty0 273--297,
  September 1995.

\bibitem[Fan et~al.(2008)Fan, Chang, Hsieh, Wang, and
  Lin]{Fan:2008:LLL:1390681.1442794}
Rong-En Fan, Kai-Wei Chang, Cho-Jui Hsieh, Xiang-Rui Wang, and Chih-Jen Lin.
\newblock \mbox{LIBLINEAR}: A library for large linear classification.
\newblock \emph{Journal of Machine Learning Research}, 9:\penalty0 1871--1874,
  June 2008.

\bibitem[Linghu and Sun(2010)]{5646323}
Bin Linghu and Bing-Yu Sun.
\newblock Constructing effective {SVM} ensembles for image classification.
\newblock In \emph{Knowledge Acquisition and Modeling (KAM), 2010 3rd
  International Symposium on}, pages 80--83, 2010.

\bibitem[List and Simon(2009)]{NL09a}
Nikolas List and Hans~Ulrich Simon.
\newblock {SVM}-optimization and steepest-descent line search.
\newblock In \emph{Proceedings of the 22nd Annual Conference on Computational
  Learning Theory}, 2009.

\bibitem[Mordelet and Vert(2011)]{citeulike:9860864}
Fantine Mordelet and Jean-Philippe~P. Vert.
\newblock {ProDiGe: Prioritization Of Disease Genes with multitask machine
  learning from positive and unlabeled examples.}
\newblock \emph{BMC bioinformatics}, 12\penalty0 (1):\penalty0 389+, 2011.

\bibitem[Osuna et~al.(1997)Osuna, Freund, and Girosi]{osuna1997}
Edgar Osuna, Robert Freund, and Federico Girosi.
\newblock {Support Vector Machines: Training and Applications}.
\newblock Technical Report AIM-1602, 1997.

\bibitem[Prokhorov(2001)]{prokhorov2001ijcnn}
Danil Prokhorov.
\newblock {IJCNN} 2001 neural network competition.
\newblock \emph{Slide presentation in {IJCNN'01}}, 2001.

\bibitem[Valentini and Dietterich(2003)]{Valentini03lowbias}
Giorgio Valentini and Thomas~G. Dietterich.
\newblock Low bias bagged support vector machines.
\newblock In \emph{In Accepted for publication, International Conference on
  Machine Learning, ICML-2003}, pages 752--759. Morgan Kaufmann, 2003.

\bibitem[Wang et~al.(2009)Wang, Mathew, Chen, Xi, Ma, and Lee]{Wang20096466}
{Shi-jin} Wang, Avin Mathew, Yan Chen, {Li-feng} Xi, Lin Ma, and Jay Lee.
\newblock Empirical analysis of support vector machine ensemble classifiers.
\newblock \emph{Expert Systems with Applications}, 36\penalty0 (3, Part
  2):\penalty0 6466 -- 6476, 2009.

\end{thebibliography}
\end{document}